\documentclass{article}

\usepackage[verbose=true,letterpaper]{geometry}

\usepackage{amsmath,amsthm,amssymb,mathtools}

\usepackage{algorithm}

\usepackage[hidelinks, hypertexnames=false, colorlinks=true, citecolor=Navy, linkcolor=Maroon, urlcolor=Orchid, bookmarksnumbered, unicode, backref=page]{hyperref}

\usepackage[noend]{algpseudocode}

\algnewcommand{\algorithmicinput}{\textbf{Input:}}
\algnewcommand{\Input}{\item[\algorithmicinput]}
\algnewcommand{\algorithmicoutput}{\textbf{Input:}}
\algnewcommand{\Output}{\item[\algorithmicoutput]}
\algnewcommand{\Break}{\textbf{break}}

\usepackage{cleveref}
\crefname{equation}{eq.}{eqs.}
\Crefname{equation}{Eq.}{Eqs.}
\crefname{algorithm}{Algorithm}{Algorithms}
\Crefname{algorithm}{Algorithm}{Algorithms}
\crefname{step}{Step}{Steps}
\Crefname{step}{Step}{Steps}
\crefname{theorem}{Theorem}{Theorems}
\Crefname{theorem}{Theorem}{Theorems}
\crefname{lemma}{Lemma}{Lemmas}
\Crefname{lemma}{Lemma}{Lemmas}
\crefname{proposition}{Proposition}{Propositions}
\Crefname{proposition}{Proposition}{Propositions}
\crefname{assumption}{Assumption}{Assumptions}
\Crefname{assumption}{Assumption}{Assumptions}
\crefname{definition}{Definition}{Definitions}
\Crefname{definition}{Definition}{Definitions}
\crefname{figure}{Figure}{Figures}
\Crefname{figure}{Figure}{Figures}
\crefname{table}{Table}{Tables}
\Crefname{table}{Table}{Tables}
\crefname{section}{Section}{Sections}
\Crefname{section}{Section}{Sections}
\crefname{appendix}{Appendix}{Appendices}
\Crefname{appendix}{Appendix}{Appendices}

\usepackage{autonum}
\makeatletter
\autonum@generatePatchedReferenceCSL{Cref}
\makeatother

\usepackage[utf8]{inputenc} 
\usepackage[T1]{fontenc}    
\usepackage{amsfonts}       
\usepackage{nicefrac}       
\usepackage{microtype}      
\usepackage{bm}             
\usepackage{bbm}

\usepackage{graphicx}
\usepackage[svgnames]{xcolor}
\usepackage{subcaption}
\usepackage{wrapfig}

\usepackage{tabularx}       
\usepackage{booktabs}       

\usepackage[round,compress]{natbib}

\usepackage{thmtools}
\usepackage{thm-restate}
\usepackage{apxproof}
\usepackage{url}            
\usepackage{braket}
\usepackage{mleftright}
\mleftright
\usepackage{multirow}
\usepackage{lipsum}
\allowdisplaybreaks
\usepackage{lscape}
\usepackage{tcolorbox}
\usepackage{xifthen}
\usepackage{xargs}
\usepackage{xspace}
\usepackage{enumerate}
\usepackage[color={red!100!green!33},colorinlistoftodos,prependcaption,textsize=small]{todonotes}

\usepackage{tikz}
\usetikzlibrary{backgrounds}
\usetikzlibrary{arrows}
\usetikzlibrary{shapes,shapes.geometric,shapes.misc}

\tikzstyle{tikzfig}=[baseline=-0.25em,scale=0.5]

\pgfkeys{/tikz/tikzit fill/.initial=0}
\pgfkeys{/tikz/tikzit draw/.initial=0}
\pgfkeys{/tikz/tikzit shape/.initial=0}
\pgfkeys{/tikz/tikzit category/.initial=0}

\pgfdeclarelayer{edgelayer}
\pgfdeclarelayer{nodelayer}
\pgfsetlayers{background,edgelayer,nodelayer,main}

\tikzstyle{none}=[inner sep=0mm]

\tikzstyle{rectangle}=[fill=white, draw=black, shape=rectangle]
\tikzstyle{circle}=[fill=white, draw=black, shape=circle]
\tikzstyle{vertex}=[fill=black, draw=none, shape=circle]
\tikzstyle{textbox}=[fill=white, draw=none, shape=rectangle]

\tikzstyle{line}=[-, fill=none]
\tikzstyle{rightarrow}=[->]
\tikzstyle{leftarrow}=[fill=none, <-]

\newcommand{\Acal}{\mathcal{A}}

\newcommand{\Dcal}{\mathcal{D}}

\newcommand{\Lcal}{\mathcal{L}}

\newcommand{\Scal}{\mathcal{S}}

\newcommand{\Xcal}{\mathcal{X}}

\newcommand{\Lhat}{\hat{L}}

\newcommand{\Ltl}{\tilde{L}}

\newcommand{\Xtl}{\tilde{X}}

\newcommand{\R}{\mathbb{R}}

\newcommand{\Z}{\mathbb{Z}}
\newcommand{\e}{\mathrm{e}}

\newcommand{\zeros}{\bm{0}}
\newcommand{\Ord}{\mathrm{O}}

\newtheorem{theorem}{Theorem}
\newtheorem{lemma}{Lemma}
\newtheorem{proposition}{Proposition}
\theoremstyle{definition}
\newtheorem{definition}{Definition}

\DeclareMathOperator*{\argmin}{argmin}

\DeclareMathOperator{\sign}{sign}
\DeclareMathOperator{\supp}{supp}

\DeclareMathOperator{\rank}{rank}
\DeclareMathOperator{\tr}{tr}

\DeclareMathOperator{\E}{\mathbb{E}}

\DeclarePairedDelimiter\abs{\lvert}{\rvert}
\DeclarePairedDelimiter\norm{\lVert}{\rVert}

\let\set\relax
\DeclarePairedDelimiter\set{\{}{\}}
\let\Set\relax
\DeclarePairedDelimiterX\Set[2]{\{}{\}}{\mspace{2mu}{#1}\;\delimsize|\;{#2}\mspace{2mu}}
\DeclarePairedDelimiter\brc{[}{]}
\DeclarePairedDelimiterX\Brc[2]{[}{]}{\mspace{2mu}{#1}\;\delimsize|\;{#2}\mspace{2mu}}
\DeclarePairedDelimiter\prn{(}{)}
\DeclarePairedDelimiterX\Prn[2]{(}{)}{\mspace{2mu}{#1}\;\delimsize|\;{#2}\mspace{2mu}}

\newif\iffigure
\figurefalse

\usepackage{svg}
\usepackage{fullpage}
\usepackage{libertine}

\title{Improved Generalization Bound and Learning of Sparsity Patterns for Data-Driven Low-Rank Approximation}

\author{%
Shinsaku Sakaue\\
The University of Tokyo\\
Tokyo, Japan\\
\href{mailto:sakaue@mist.i.u-tokyo.ac.jp}{sakaue@mist.i.u-tokyo.ac.jp}
\and
Taihei Oki\\
The University of Tokyo\\
Tokyo, Japan\\
\href{mailto:oki@mist.i.u-tokyo.ac.jp}{oki@mist.i.u-tokyo.ac.jp}
}
\date{}

\begin{document}

\maketitle

\begin{abstract}
  Learning sketching matrices for fast and accurate low-rank approximation (LRA) has gained increasing attention. Recently, Bartlett, Indyk, and Wagner (COLT 2022) presented a generalization bound for the learning-based LRA. Specifically, for rank-$k$ approximation using an $m \times n$ learned sketching matrix with $s$ non-zeros in each column, they proved an $\tilde{\mathrm{O}}(nsm)$ bound on the \emph{fat shattering dimension} ($\tilde{\mathrm{O}}$ hides logarithmic factors). We build on their work and make two contributions.
  \begin{enumerate}
    \item We present a better $\tilde{\mathrm{O}}(nsk)$ bound ($k \le m$). En route to obtaining this result, we give a low-complexity \emph{Goldberg--Jerrum algorithm} for computing pseudo-inverse matrices, which would be of independent interest.
    \item We alleviate an assumption of the previous study that sketching matrices have a fixed sparsity pattern. We prove that learning positions of non-zeros increases the fat shattering dimension only by ${\mathrm{O}}(ns\log n)$. In addition, experiments confirm the practical benefit of learning sparsity patterns.
  \end{enumerate}
\end{abstract}

\newcommand{\Ibb}{\mathbb{I}}
\newcommand{\pdim}{\mathrm{pdim}}
\newcommand{\FSdim}{\mathrm{fatdim}}
\newcommand{\eFSdim}{\mathrm{fatdim}_\varepsilon}
\newcommand{\VCdim}{\mathrm{VCdim}}

\newcommand{\svd}{\mathrm{SVD}}
\newcommand{\scw}{\mathrm{SCW}}
\newcommand{\train}{\text{train}}
\newcommand{\test}{\text{test}}
\newcommand{\pcomp}{p}
\newcommand{\nparam}{\ell}

\newcommand{\fix}{\textbf{Fix}\xspace}
\newcommand{\learn}{\textbf{Learn}\xspace}
\newcommand{\dense}{\textbf{Dense}\xspace}

\section{INTRODUCTION}\label{sec:introduction}
Low-rank approximation (LRA) has played a crucial role in analyzing matrix data.
Although the singular value decomposition (SVD) provides an optimal LRA, it is too costly when the data size is huge.
To overcome this limitation, researchers have developed fast LRA methods with \emph{sketching}, whose basic form is as follows: given an input matrix $A \in \R^{n \times d}$ and a target low rank $k$, choose a sketching matrix $S \in \R^{m \times n}$ with $k \le m \le \min\set{n, d}$ and compute an LRA matrix for $SA \in \R^{m \times d}$.
If $S$ is drawn from an appropriate distribution, the resulting matrix is a good LRA of $A$ with high probability \citep{Sarlos2006-ob,Clarkson2009-yb,Clarkson2017-jq}.
This randomized sketching paradigm has led to various time- and space-efficient algorithms in numerical linear algebra.
We refer the reader to \citep{Woodruff2014-ya,Martinsson2020-mn} for more details of this area.

While such LRA methods with randomized sketching enjoy rigorous guarantees even for worst-case input matrices, a recent line of work \citep{Indyk2019-cn,Liu2020-hu,Indyk2021-yn} suggests that \emph{learning-based} LRA methods can attain significantly smaller approximation errors when we can use past data to better handle future data.
They have achieved fast and more accurate LRA by learning sketching matrices $S$ to minimize approximation errors over past data.

As for the theoretical side of learning-based LRA, \citet{Bartlett2022-mu} recently presented generalization bounds for learning sketching matrices.
Specifically, they proved an $\tilde\Ord(nsm)$\footnote{We use $\tilde\Ord$ and $\tilde\Omega$ to hide logarithmic factors.} upper bound on the \emph{fat shattering dimension} for learning an $m \times n$ sketching matrix with $s$ non-zeros at \emph{fixed} positions in each column.
They also showed an $\Omega(ns)$ lower bound.
We give an overview of their work in \cref{subsec:bartlett-overview}.

Their study has raised some natural questions.
For example, can we narrow the $\tilde\Ord(m)$ gap between the upper and lower bounds?
Moreover, generalization bounds for learning-based LRA with \emph{changeable} sparsity patterns are awaited since learning positions of non-zeros is considered to be a promising direction \citep{Indyk2021-yn} and its effectiveness has been partly confirmed \citep{Liu2020-hu}.

\subsection{Our Contribution}
Building on \citep{Bartlett2022-mu}, we address the aforementioned questions and make two contributions.

First, we improve the previous $\tilde\Ord(nsm)$ upper bound by replacing the $\Ord(m)$ factor with $\Ord(\log m)$, which leads to a better $\tilde\Ord(nsk)$ bound ($k \le m$).
Although the \emph{sketching dimension}, $m$, is often set to, for example, $4k$ in practice, there is no such theoretical relation as $m = \Ord(k)$.
Thus, our bound indeed improves the previous one.
We take the same proof strategy as \citep{Bartlett2022-mu} and represent computational procedures of a loss function by a \emph{Goldberg--Jerrum (GJ) algorithm}. Our technical contribution is to develop a new GJ algorithm for computing pseudo-inverse matrices with a smaller \emph{predicate complexity} than the previous one, which would be of independent interest. To demonstrate its usefulness, we also give a generalization bound for a learning-based Nystr\"om method by using our GJ algorithm.

Second, we give a generalization bound for learning-based LRA with changeable sparsity patterns. Supposing we can learn both positions and values of $ns$ non-zeros in a sketching matrix $S$, we prove that our $\tilde\Ord(nsk)$ upper bound on the fat shattering dimension increases only by $\Ord(ns\log n)$, despite the presence of exponentially many possible sparsity patterns in $ns$.
Hence, the bound remains $\tilde\Ord(nsk)$ (ignoring $\Ord(\log n)$) even when the sparsity pattern can change.
Also, experiments show that a recent efficient learning-based LRA method \citep{Indyk2021-yn}, which used fixed sparsity patterns, can achieve higher accuracy with changeable sparsity patterns, suggesting the practical benefit of our result.

\subsection{Related Work}
The most relevant study to ours is \citep{Bartlett2022-mu}.
They proved generalization bounds for learning-based LRA and other methods in numerical linear algebra.
Other theoretical results related to learning-based LRA include \emph{safeguard} guarantees \citep{Indyk2019-cn} and \emph{consistency} \citep{Indyk2021-yn}, which are different from generalization guarantees, as mentioned in \citep[Section 2.5]{Bartlett2022-mu}.

\citet{Gupta2017-ng} initiated the study of a PAC-learning approach to algorithm configuration, which is also called \emph{data-driven algorithm design} \citep{Balcan2021-fy}.
Recent studies have presented generalization bounds for various learning-based algorithms, e.g., integer programming methods \citep{Balcan2018-pe,Balcan2021-kv,Balcan2022-em}, clustering \citep{Balcan2020-im}, and heuristic search \citep{Sakaue2022-nt}.
\citet{Balcan2021-jv} presented a general theory for deriving generalization bounds based on piecewise structures of \emph{dual} function classes.
Their idea, however, does not lead to strong guarantees in learning-based LRA, as discussed in \citep[Appendix E]{Bartlett2022-mu}.
As with \citep{Bartlett2022-mu}, we consider a class of \emph{proxy loss} functions to obtain a generalization bound.
This idea has a slight connection to \citep{Balcan2020-gm}, which approximates dual functions with simpler ones, while the technical details are different.

\section{BACKGROUND}
For any positive integer $n$, let $[n] = \set*{1,\dots,n}$.
Let $\sign(\cdot)$ be the sign function that takes $x \in \R$ as input and returns $+1$ if $x>0$, $-1$ if $x<0$, or $0$ if $x=0$.
We define the degree of a polynomial by its total degree.
The degree of a rational function refers to the maximum of its numerator's and denominator's degrees, where the fraction is reduced to the lowest terms.

Let $\rank$, $\tr$, and $\det$ denote the rank, trace, and determinant.
Let $\norm{A}_F = \sqrt{\tr(A^\top A)}$ denote the Frobenius norm of a matrix $A$.
The Moore--Penrose pseudo-inverse of a matrix $A$ is denoted by $A^\dagger$.
SVD refers to the compact singular value decomposition, i.e., for $A \in \R^{n \times d}$ with $\rank(A) = r$, SVD computes $U\in \R^{n \times r}$, $\Sigma \in \R^{r \times r}$, and $V \in \R^{d \times r}$ with $A = U \Sigma V^\top$.
For any vector $x \in \R^n$, let $\supp(x) \subseteq [n]$ denote the set of indices of non-zeros.
A \emph{sparsity pattern}, $J \subseteq [n]$, of $x$ indicates that $x_i$ is \emph{allowed} to be non-zero if and only if $i \in J$, hence $\supp(x) \subseteq J$.

\subsection{Learning Theory}\label{subsec:learning-theory}
Let $\Xcal$ be a domain of inputs, $\Dcal$ a distribution over $\Xcal$, and $\Lcal \subseteq [0,1]^\Xcal$ a class of loss functions.
In our case, $\Xcal$ is a class of input matrices, and each $L \in \Lcal$ measures the approximation error of LRA and is parametrized by a sketching matrix (see \cref{subsec:bartlett-overview}).
For $\delta \in (0,1)$ and $\varepsilon > 0$,
we say $\Lcal$ admits $(\varepsilon, \delta)$-\emph{uniform convergence} with $N$ samples if for i.i.d.\ draws $\Xtl = \set{x_1,\dots,x_N} \sim \Dcal^N$, it holds that
\[
  \Pr_{\Xtl}
  \brc*{
    \forall L \in \Lcal, \,
    \abs*{\frac{1}{N}\sum_{i=1}^N L(x_i) - \mathop{\E}_{x\sim\Dcal}[L(x)]} \le \varepsilon
  }
  \ge 1-\delta.
\]
If such a uniform bound over $\Lcal$ holds, we can bound the gap between the empirical and expected losses regardless of how sketching matrices are learned (e.g., manual or automatic).

The following \emph{pseudo-} and \emph{fat shattering dimensions} are fundamental notions of the complexity of function classes.
\begin{definition}[Pseudo- and fat shattering dimensions]\label{def:dim}
	Let $\Lcal \subseteq [0,1]^\Xcal$ be a class of functions.
	We say an input set $\set*{x_1,\dots,x_N}\subseteq \Xcal$ is \textit{(pseudo) shattered} by $\Lcal$ if there exist threshold values, $t_1,\dots,t_N \in \R$, satisfying the following condition:
  for every $I \subseteq [N]$, there exists $L \in \Lcal$ such that
	\begin{equation}\label{eq:shatter}
    i \in I \Leftrightarrow L(x_i) > t_i.
  \end{equation}
  For $\gamma > 0$, we say $\set*{x_1,\dots,x_N}\subseteq \Xcal$ is \textit{$\gamma$-fat shattered} by $\Lcal$ if the above condition holds with replacement of \eqref{eq:shatter} by
  \begin{equation}
    i \in I \Rightarrow L(x_i) > t_i + \gamma
    \quad \text{and} \quad
    i \notin I \Rightarrow L(x_i) < t_i - \gamma.
  \end{equation}
  The \emph{pseudo-dimension}, $\pdim(\Lcal)$, and \emph{$\gamma$-fat shattering dimension}, $\FSdim_\gamma(\Lcal)$, are the maximum size of a set that is pseudo and $\gamma$-fat shattered, respectively, by $\Lcal$.
\end{definition}
It is well-known that $N = \Omega(\varepsilon^{-2}\cdot(\pdim(\Lcal) + \log\delta^{-1}))$ samples are sufficient for ensuring $(\varepsilon, \delta)$-uniform convergence, and a similar guarantee holds if $\FSdim_\gamma(\Lcal)$ with $\gamma = \Omega(\varepsilon)$ is bounded.
We refer the reader to \citep[Theorems 19.1 and 19.2]{Anthony1999-mm} for details.

\subsection{Low-Rank Approximation}\label{subsec:lra}
Let $A \in \R^{n\times d} $ be an input matrix with $n \ge d$.
We assume $\rank(A) > 0$ and $\norm{A}_F^2 = 1$ by normalization.
For $k \in [d]$, we consider computing a rank-$k$ approximation of $A$.
Let $\brc{A}_k \in \R^{n \times d}$ denote an optimal rank-$k$ approximation, i.e.,
\[
  \brc{A}_k \in \argmin\Set*{\norm{A - X}_F^2}{X \in \R^{n \times d}, \rank(X) = k}.
\]
Although we can compute $\brc{A}_k$ with SVD in $\Ord(nd^2)$ time \citep[Section 8.6.3]{Golub2013-xr}, this approach is time and space consuming when $A$ is huge.

\Cref{alg:scw} presents an efficient LRA algorithm with a sketching matrix $S \in \R^{m \times n}$ \citep{Sarlos2006-ob,Clarkson2009-yb,Clarkson2017-jq}, which is called the SCW algorithm after the authors' acronyms.
\Cref{alg:scw} is more efficient than computing $\brc{A}_k$ if we set the sketching dimension, $m$, to a much smaller value than $d$, whereas we need $m \ge k$ to get a rank-$k$ approximation.
Let $\scw_k(S, A)$ denote the output of \cref{alg:scw} with a sketching matrix $S$ and an input matrix $A$.
It is known that for $\alpha > 0$, sketching matrices with $m = \tilde\Omega(k/\alpha)$ drawn from an appropriate distribution satisfy $\norm{A - \scw_k(S, A)}_F \le (1+\alpha)\norm{A - \brc*{A}_k}_F$ with high probability (e.g., \citep[Section 4.1]{Woodruff2014-ya}).

\citet{Indyk2019-cn} showed that machine-learned sketching matrices often enable more accurate LRA than random ones in practice.
Given a training dataset $\Acal_{\train} \subseteq \R^{n \times d}$ of input matrices, they proposed to learn $S$ by minimizing the empirical risk $\frac{1}{|\Acal_{\train}|}\sum_{A\in\Acal_{\train}} \norm{A - \scw_k(S, A)}_F^2$.
Specifically, they learned sparse $S$ with the stochastic gradient descent method (SGD) by regarding non-zeros in $S$ at fixed positions as tunable parameters (where the sparsity of $S$ makes $\scw_k$ efficient).
Later, researchers further studied learning-based LRA methods \citep{Liu2020-hu,Ailon2021-ia,Indyk2021-yn}, which we will overview in \cref{subsec:experiment-background}.

\begin{algorithm}[tb]
	\caption{$\scw_k(S, A)$}
	\label{alg:scw}
	\begin{algorithmic}[1]
    \State Compute $SA$
    \If{$SA$ is a zero matrix}
    \State \Return an $n\times d$ zero matrix
    \EndIf
		\State $U$, $\Sigma$, $V$ $\gets \svd(SA)$ \Comment{$SA = U\Sigma V^\top$}
    \State Compute $AV$
    \State \Return $\brc{AV}_k V^\top$
	\end{algorithmic}
\end{algorithm}

\subsection{Overview of \texorpdfstring{\citep{Bartlett2022-mu}}{Bartlett et al., 2022}}\label{subsec:bartlett-overview}
\citet{Bartlett2022-mu} formally studied learning-based LRA as a statistical learning problem.
Let $\Acal \subseteq \R^{n\times d}$ be a class of input matrices and $\Scal\subseteq \R^{m\times n}$ a class of sketching matrices, where every $S \in \Scal$ has up to $s$ non-zeros in each column and the sparsity pattern is identical for all $S \in \Scal$.
Define a loss function $L:\Scal\times\Acal\to[0, 1]$\footnote{$L(S, A)$ is at most $\norm{A}_F^2 = 1$, as in \citep{Bartlett2022-mu}.} based on $\scw_k$ as
\begin{equation}\label{eq:scw-loss}
  L(S, A) = \norm{A - \scw_k(S, A)}_F^2.
\end{equation}
Let $\Lcal = \set{L(S, \cdot)}_{S \in \Scal} \subseteq {[0, 1]}^\Acal$ be the class of loss functions, where each $L(S, \cdot) \in \Lcal$ is specified by $ns$ tunable parameters (non-zeros of $S$) and measures the approximation error of $\scw_k(S, \cdot)$.
The authors presented the following $\tilde\Ord(nsm)$ bound on the $\varepsilon$-fat shattering dimension of $\Lcal$.
\begin{theorem}[{\citet[Theorem 2.2]{Bartlett2022-mu}}]\label{thm:bartlett-fatdim}
  For sufficiently small $\varepsilon > 0$, the $\varepsilon$-fat shattering dimension of $\Lcal$ is bounded as
  \[
    \eFSdim(\Lcal) = \Ord(ns \cdot (m + k\log(d/k) + \log(1/\varepsilon))).
  \]
\end{theorem}

Intuitively, we can bound $\eFSdim(\Lcal)$ by assessing the complexity of computational procedures for evaluating $L(S, A)$.
In the LRA setting, however, directly bounding $\eFSdim(\Lcal)$ is not easy since $\scw_k$ makes black-box use of SVD.
The authors have overcome this difficulty by considering a class $\hat\Lcal_\varepsilon$ of appropriate \emph{proxy loss} functions, which we can evaluate with relatively simple computational procedures, and by bounding its pseudo-dimension, $\pdim(\hat\Lcal_\varepsilon)$.
As in the following definition, each $\Lhat_\varepsilon(S, \cdot) \in \hat\Lcal_\varepsilon$ is evaluated with a \emph{power-method}-based procedure so that $\Lhat_\varepsilon(S, A)$ gives a sufficiently accurate approximation of $L(S, A)$.

\begin{definition}[Proxy loss]\label{def:proxy}
  For any $A \in \Acal$, $S \in \Scal$, and $\varepsilon>0$, the proxy loss $\Lhat_\varepsilon(S, A)$ is computed as follows:
  \begin{enumerate}
    \item Compute $B = A{(SA)}^\dagger(SA)$.{\label[step]{item:pinv-comp}}
    \item For all possible $P_i \in \R^{d \times k}$ ($i = 1,\dots,\binom{d}{k}$) whose columns are $k$ distinct standard vectors in $\R^d$, compute $Z_i = {(BB^\top)}^qBP_i$, where $q = \Ord(\varepsilon^{-1}\log(d/\varepsilon))$.{\label[step]{item:for-all-Pi}}
    \item  Choose $Z = Z_i$ that minimizes $\norm{B - Z_iZ_i^\dagger B}_F^2$.{\label[step]{item:choose-Z}}
    \item $\Lhat_\varepsilon(S, A) = \norm{A - ZZ^\dagger B}_F^2$.
  \end{enumerate}
  Given the class $\Scal$ of sketching matrices, the class of proxy loss functions is defined as $\hat\Lcal_\varepsilon = \set{\Lhat_\varepsilon(S, \cdot)}_{S \in \Scal}$.
\end{definition}

As discussed in \citep[Section 5.3]{Bartlett2022-mu}, it holds that $\eFSdim(\Lcal) \le \pdim(\hat\Lcal_\varepsilon)$.
Therefore, an upper bound on $\pdim(\hat\Lcal_\varepsilon)$ immediately implies that on $\eFSdim(\Lcal)$.

A benefit of considering $\hat\Lcal_\varepsilon$ is that analyzing its complexity is easier than $\Lcal$.
The authors upper bounded $\pdim(\hat\Lcal_\varepsilon)$ by modeling the computational procedure of $\Lhat_\varepsilon$ as a \emph{Goldberg--Jerrum algorithm} \citep{Goldberg1995-af}.\footnote{Such a notion is often called the \emph{algorithmic computation tree}. Still, we here call it a GJ algorithm to be consistent with \citep{Bartlett2022-mu}. Although their original definition does not contain the equality condition in branch nodes, dealing with equalities is easy due to \citep[Corollary 2.1]{Goldberg1995-af}.}

\begin{definition}[Goldberg--Jerrum algorithm]\label{def:gj}
  A GJ algorithm $\Gamma$ takes real values as input, and its procedure is represented by a binary tree with the following two types of nodes:
  \begin{itemize}
    \item Computation node that executes an arithmetic operation $v^{\prime \prime} = v \odot v^\prime$, where $\odot \in \set*{+, -, \times, \div}$.
    \item Branch node with an out-degree of $2$, where branching is specified by the evaluation of a condition of the form $v \ge 0$ ($v \le 0$) or $v = 0$.
  \end{itemize}
  In both cases, $v$ and $v^\prime$ are either inputs or values computed at ancestor nodes.
  Once input values are given, $\Gamma$ proceeds along a root--leaf path on the tree and sequentially performs operations specified by nodes on the path.
\end{definition}

Then, they defined two notions, the \emph{degree} and \emph{predicate complexity}, to measure the complexity of GJ algorithms.
\begin{definition}[Degree and predicate complexity]
  The \emph{degree} of a GJ algorithm is the maximum degree of any rational function of input variables it computes.
  The \emph{predicate complexity} of a GJ algorithm is the number of distinct rational functions that appear at its branch nodes.
  If a GJ algorithm has the degree and predicate complexity of at most $\Delta$ and $\pcomp$, respectively, we call it a $(\Delta, \pcomp)$-GJ algorithm.
\end{definition}

The following theorem says that if we can check whether a loss function value exceeds a threshold value or not using a $(\Delta, \pcomp)$-GJ algorithm with small $\Delta$ and $\pcomp$, the class of such loss functions has a small pseudo-dimension.

\begin{theorem}[{\citet[Theorem 3.3]{Bartlett2022-mu}}]\label{thm:bartlett-main}
  Let $\Xcal$ be an input domain and $\Lcal = \Set{L_\rho:\Xcal\to\R}{\rho \in \R^\nu}$ a class of functions parameterized by $\rho \in \R^\nu$.
  Assume that for every $x \in \Xcal$ and $t \in \R$, there is a $(\Delta, \pcomp)$-GJ algorithm $\Gamma_{x,t}$ that takes $\rho \in \R^\nu$ as input and returns ``true'' if $L_\rho(x) > t$ and ``false'' otherwise.
  Then, it holds that
  \[
    \pdim(\Lcal) = \Ord(\nu \log(\pcomp\Delta)).
  \]
\end{theorem}

The authors proved that for any $A \in \Acal$ and $t \in \R$, whether $\Lhat_\varepsilon(S, A) > t$ or not can be checked by a $(\Delta, \pcomp)$-GJ algorithm $\Gamma_{A, t}$ with $\Delta = \Ord(mk\varepsilon^{-1}\log(d/\varepsilon))$ and
\begin{equation}\label{eq:kappa-factors}
  \pcomp = 2^m\cdot 2^{\Ord(k)}\cdot {(d/k)}^{3k},
\end{equation}
where input variables are $ns$ non-zeros of $S$, i.e., $\nu = ns$.
Therefore, \cref{thm:bartlett-main} implies
\begin{equation}\label{eq:bartlett-pdim-upper}
  \pdim(\hat\Lcal_\varepsilon) = \Ord(ns \cdot (m + k\log(d/k) + \log(1/\varepsilon))).
\end{equation}
The same bound applies to $\eFSdim(\Lcal)$ ($\le \pdim(\hat\Lcal_\varepsilon)$), obtaining \cref{thm:bartlett-fatdim}.
They also gave an $\Omega(ns)$ lower bound on $\eFSdim(\Lcal)$; hence it is tight up to an $\tilde\Ord(m)$ factor.

\subsection{Warren's Theorem}\label{subsec:warren}
Warren's theorem \citep{Warren1968-hp} is a useful tool to evaluate the complexity of a class of polynomials.
The following extended version that allows the sign to be zero is presented in \citep[Corollary 2.1]{Goldberg1995-af}.
\begin{theorem}[Warren's theorem]\label{thm:warren}
  Let $\set*{f_1,\dots,f_N}$ be a set of $N$ polynomials of degree at most $\Delta$ in $\nu$ real variables $\rho \in \R^\nu$.
  If $N \ge \nu$, there are at most ${(8\e N\Delta/\nu)}^\nu$ distinct tuples of $(\sign(f_1(\rho)),\dots,\sign(f_N(\rho))) \in \set{-1, 0, +1}^N$.
\end{theorem}

This theorem is a key to proving \cref{thm:bartlett-main}, and we will also use it in \cref{sec:sparsity-pattern}.
To familiarize ourselves with the theorem, we give a proof sketch of \cref{thm:bartlett-main}. From the statement assumption in \cref{thm:bartlett-main}, whether $L_\rho(x) > t$ is determined by sign patterns of $\pcomp$ polynomials of degree at most $\Delta$ in $\rho\in\R^\nu$ that appear at the branch nodes of the GJ algorithm, $\Gamma_{x, t}$.
Thus, when $x_1,\dots,x_N \in \Xcal$ and $t_1,\dots,t_N \in \R$ are given, the number of distinct outcomes (or tuples of $N$ Booleans) of GJ algorithms $\Gamma_{x_1, t_1}, \dots, \Gamma_{x_N, t_N}$, which take common $\rho$ as input, is bounded by the number of all possible sign patterns of $N\pcomp$ polynomials of degree at most $\Delta$ in $\rho$.
From Warren's theorem, the number of such sign patterns is at most ${(8\e N\pcomp\Delta/\nu)}^\nu$, which must be at least $2^N$ to shatter $x_1,\dots,x_N$.
The largest $N$ with ${(8\e N\pcomp\Delta/\nu)}^\nu \ge 2^N$ gives the $\Ord(\nu \log(\pcomp\Delta))$ bound on $\pdim(\Lcal)$, as in \cref{thm:bartlett-main}.

\section{IMPROVED UPPER BOUND}\label{sec:improved-upper}
We obtain an $\tilde\Ord(nsk)$ bound on $\eFSdim(\Lcal)$ by replacing the $\Ord(m)$ factor in \eqref{eq:bartlett-pdim-upper} with $\Ord(\log m)$.
To this end, we reduce the $2^m$ factor in the predicate complexity \eqref{eq:kappa-factors} to $m$.

Note that, although concatenating random matrices with $m = \tilde\Ord(k/\alpha)$ rows guarantees the $(1+\alpha)$-approximation as mentioned in \cref{subsec:lra} (known as safeguard guarantees), our improvement is not meaningless since $m$ can be much lager than $k$.
For example, even if we admit errors of $\scw_k$ relative to $\brc{A}_k$ to the magnitude of $\alpha \approx \varepsilon$, $m = \tilde\Ord(k/\alpha) \simeq \tilde\Ord(k/\varepsilon)$ does not imply $m = \Ord(k \log (d/k) + \log(1/\varepsilon))$, hence $\tilde\Ord(nsk)$ can be significantly smaller than $\tilde\Ord(nsm)$.


\subsection{Previous Approach}\label{subsec:pinv-previous}
We first explain where the $2^m$ factor comes from in \citep{Bartlett2022-mu}.
By carefully expanding the proof of \citep[Lemma 5.6]{Bartlett2022-mu}, one can confirm that it is caused by Step~\ref{item:pinv-comp} in \cref{def:proxy}, where a GJ algorithm computes $A(SA)^\dagger(SA)$.
For this step, they used an $(\Ord(m), 2^m)$-GJ algorithm that computes $Z^\dagger Z$ for an input matrix $Z$ with $m$ rows \citep[Lemma 5.2]{Bartlett2022-mu}. We below describe their GJ algorithm for later convenience.
In what follows, let $I_r$ denote the $r \times r$ identity matrix for any $r \in \Z_{>0}$.

An essential tool for obtaining the GJ algorithm is the matrix inversion formula by the Cayley--Hamilton theorem.\footnote{\citet{Bartlett2022-mu} alternatively used a recursive formula of \citep{Csanky1976-kp}. This difference does not affect the conclusion.}

\begin{proposition}\label{prop:csanky}
  Let $M$ be an $r \times r$ real matrix and
  \[
    \det(\lambda I_r - M) = \lambda^r + c_1\lambda^{r-1} + \cdots + c_r
  \]
  the characteristic polynomial of $M$.
  If $M$ is invertible, we have $c_r = {(-1)}^r \det(M) \neq 0$ and
  \[
    M^{-1} = -\frac{1}{c_r} \cdot (M^{r-1} + c_1 \cdot M^{r-2} + \dots + c_{r-1}\cdot I_r).
  \]
\end{proposition}

Let $Z$ be an input matrix with $m$ rows of rank $r \le m$.
Their GJ algorithm computes $Z^\dagger Z$ as follows.
It first finds a matrix $Y$ with $r$ linearly independent rows selected from the rows of $Z$.
Since $Y$ spans the row space of $Z$, it holds $Z^\dagger Z = Y^\top{(YY^\top)}^{-1}Y$; their algorithm computes this using \cref{prop:csanky} with $M = YY^\top$.
Note that we have
\begin{align}
  c_i = {(-1)}^i \sum_{S: |S| = i} \det M[S]
  & &
  \text{for $i = 1,\dots,r$,}
\end{align}
where $M[S]$ is the principal minor of $M$ with indices $S \subseteq [r]$;
hence, if we take entries of $M$ to be variables, $c_1,\dots,c_r$ are polynomials of degree at most $r$.
Thus, regarding entries of $Z$ as variables, every rational function that appears in the above procedure has a degree of $\Ord(m)$.

What remains to be discussed is how to find $Y$ of full row rank.
To achieve this, their GJ algorithm goes over the rows of $Z$ and sequentially adds appropriate rows to $Y$ in a greedy fashion.
Whenever adding a new row, it checks whether the resulting $Y$ has full row rank by examining whether $\det(YY^\top) \neq 0$ or not.
This procedure involves polynomials of degree $\Ord(m)$, and the number of branch nodes is up to $2^m$ depending on which rows of $Z$ are selected, resulting in the $2^m$ predicate complexity.

\subsection{Our Result}\label{subsec:pinv-ours}
We present an $(\Ord(m), m)$-GJ algorithm for computing $Z^\dagger$ (right-multiplying $Z$ only increases the degree by one).

\begin{lemma}\label{lem:better-gj-alg}
  Let $Z$ be an input matrix with $m$ rows.
  There is an $(\Ord(m), m)$-GJ algorithm that computes $Z^\dagger$.
\end{lemma}

Our key idea is to begin by determining $r = \rank(ZZ^\top)$ with $m$ branch nodes, instead of branching to determine the choice of rows of $Z$.
Once $r$ is fixed, we can calculate $Z^\dagger$ without branching by the following formula.

\begin{proposition}[{\citet[Theorem 3]{Decell1965-iz}}]\label{prop:pinv}
  Let $Z$ be a matrix with $m$ rows and $c_1,\dots,c_m$ the coefficients of the characteristic polynomial of $M = ZZ^\top \in \R^{m \times m}$, i.e.,
  \[
    \det(\lambda I_m - M) = \lambda^m + c_1\lambda^{m-1} +\dots+ c_m.
  \]
  If $r \ge 1$ is the largest index with $c_r\neq 0$, we have
\[
  Z^\dagger = -\frac{1}{c_r} \cdot Z^\top
  \prn*{
    M^{r-1} + c_1 \cdot M^{r-2} +\dots+ c_{r-1} \cdot I_m
  }.
\]
If $c_1 = \dots = c_m = 0$, $Z^\dagger$ is a zero matrix.
\end{proposition}

By using this formula in lieu of \cref{prop:csanky}, we can obtain an $(\Ord(m), m)$-GJ algorithm that computes $Z^\dagger Z$.

\begin{proof}[Proof of \cref{lem:better-gj-alg}]
  We give a concrete GJ algorithm.
  Let $M = ZZ^\top$.
  First, we compute the coefficients $c_1,\dots,c_m$ of $\det(\lambda I_m - M)$, which are polynomials of degree $\Ord(m)$ in the entries of $Z$.
  Then, check whether $c_i \neq 0$ in decreasing order of $i$.
  Once we find $c_i \neq 0$, set $r = i$ as the largest index $r$ with $c_r \neq 0$.
  Note that this requires only $m$ branch nodes.
  If $c_m = \dots = c_1 = 0$, let $Z^\dagger$ be a zero matrix.
  Otherwise, we compute $Z^\dagger$ as in \cref{prop:pinv}.
  Every rational function in the above calculation has a degree of $\Ord(m)$ in $Z$.
  Thus, we obtain a desired $(\Ord(m),m)$-GJ algorithm.
\end{proof}

By performing Step~\ref{item:pinv-comp} in \cref{def:proxy} with our GJ algorithm, we can replace the $\Ord(m)$ factor in the upper bound \eqref{eq:bartlett-pdim-upper} with $\Ord(\log m)$, thus improving \cref{thm:bartlett-fatdim} as follows.
\begin{proposition}\label{prop:fat-dim-improved}
  For sufficiently small $\varepsilon > 0$, the $\varepsilon$-fat shattering dimension of $\Lcal$ is bounded as
  \[
    \eFSdim(\Lcal) = \Ord(ns \cdot (\log m + k\log(d/k) + \log(1/\varepsilon))).
  \]
\end{proposition}

\subsection{Application to the Nystr\"om Method}\label{subsec:nystrom}
We briefly digress to demonstrate the usefulness of our GJ algorithm (\cref{lem:better-gj-alg}).
We here consider the classical Nystr\"om method \citep{Nystrom1930-dx}.
The method takes a positive semidefinite matrix $A \in \R^{n \times n}$ as input and computes its rank-$r$ approximation as $AS{(S^\top AS)}^\dagger {(AS)}^\top$, where $S \in \R^{n \times r}$ is a sketching matrix.
Unlike the SCW algorithm (\cref{alg:scw}), it does not involve SVD, hence more efficient.
Thus, it is a popular choice when handling large Laplacian and kernel matrices \citep{Gittens2016-pp}.

As with learning-based LRA methods discussed so far, we can naturally combine the Nystr\"om method with learning of sketching matrices.
Specifically, defining a loss function as
\begin{equation}\label{eq:nystrom-loss}
  L(S, A) = \norm{A - AS{(S^\top AS)}^\dagger {(AS)}^\top}_F^2,
\end{equation}
we can learn high-performing sketching matrices from past data of $A$ by minimizing the empirical risk.
When it comes to generalization guarantees, we are interested in the pseudo-dimension of $\Lcal = \set{L(S, \cdot)}_{S \in \Scal}$ with $L$ defined as in \eqref{eq:nystrom-loss}, where we let $\Scal \subseteq \R^{n \times r}$ be a class of sketching matrices with $\nu$ non-zeros at fixed positions.

We analyze the pseudo-dimension of $\Lcal$ by modeling the computational procedure of $L(S, A)$ defined in \eqref{eq:nystrom-loss} as a GJ algorithm.
We first compute ${(S^\top A S)}^\dagger$ with our GJ algorithm (\cref{lem:better-gj-alg}), whose degree and predicate complexity are $\Ord(r)$ and $r$, respectively, where entries of $S$ are variables.
Other operations for computing $L(S, A)$ require no branch nodes, and the degree remains $\Ord(r)$.
Consequently, we can compute $L(S, A)$ with an $(\Ord(r), r)$-GJ algorithm, and thus \cref{thm:bartlett-main} implies the following bound on $\pdim(\Lcal)$.
\begin{proposition}
  For the class of $\Lcal$ of loss functions \eqref{eq:nystrom-loss}, each of which is parameterized by an $n \times r$ sketching matrix $S \in \Scal$ with $\nu$ non-zeros at fixed positions, it holds that
  \[
    \pdim(\Lcal) = \Ord(\nu\log r).
  \]
\end{proposition}
We can also deal with changeable sparsity patterns by using \cref{thm:sparse-main}, which we will show in \cref{sec:sparsity-pattern}.
In this case, it will immediately follow that $\pdim(\Lcal) = \Ord(\nu\log(nr))$.

Note that if we compute ${(S^\top A S)}^\dagger$ with the previous GJ algorithm described in \cref{subsec:pinv-previous}, its predicate complexity is $2^r$, resulting in $\pdim(\Lcal) = \Ord(\nu r\log r)$.
Thus, this example suggests that our GJ algorithm can yield much better generalization bounds for classes of functions involving pseudo-inverse computation.

\section{LEARNING SPARSITY PATTERNS}\label{sec:sparsity-pattern}
This section studies generalization bounds when sparsity patterns of sketching matrices can change.
We show that even if the class $\Scal$ of sketching matrices contains all $m \times n$ matrices with $ns$ non-zeros, the fat shattering dimension of $\Lcal = \set{L(S, \cdot)}_{S \in \Scal}$ increases only by $\Ord(ns\log n)$.

\subsection{General Result}
To deal with changeable sparsity patterns, we first present an extended version of \cref{thm:bartlett-main}.

\begin{theorem}\label{thm:sparse-main}
  Let $\Xcal$ be an input domain and $\Lcal \subseteq \R^\Xcal$ a class of functions with $\nparam $ parameters $\rho \in \R^\nparam $ that is $\nu$-sparse, i.e.,
  \[
    \Lcal = \Set{L_\rho:\Xcal\to\R}{\rho \in \R^\nparam , |\supp(\rho)| \le \nu}.
  \]
  Assume that for every $x \in \Xcal$ and $t \in \R$, there is a $(\Delta, \pcomp)$-GJ algorithm, $\Gamma_{x,t}$, that takes a $\nu$-sparse variable vector $\rho \in \R^\nparam $ as input and returns ``true'' if $L_\rho(x) > t$ and ``false'' otherwise.
  Then, we have
  \[
    \pdim(\Lcal) = \Ord(\nu\log(\nparam \pcomp\Delta)).
  \]
\end{theorem}
Compared with \cref{thm:bartlett-main}, there are $\nparam $ ($\ge \nu$) parameters, which are restricted to be $\nu$-sparse.
If we naively use \cref{thm:bartlett-main} without taking the sparsity into account, the pseudo-dimension bound turns out $\tilde\Ord(\nparam )$, even though every $L_\rho$ has only $\nu$ tunable non-zero parameters.
Our \cref{thm:sparse-main} provides a refined bound that grows only logarithmically with $\nparam $ and keeps the linear dependence on $\nu$.

The following proof idea comes from a PAC approach to one-bit compressed sensing \citep{Ahsen2019-yj}, but how to use the idea is significantly different; indeed, the previous study does not combine it with Warren's theorem.

\begin{proof}[Proof of \cref{thm:sparse-main}]
  The proof proceeds similarly to that of \citep[Theorem 3.3]{Bartlett2022-mu} (sketched in \cref{subsec:warren}), but we must take changeable sparsity patterns into account.

  We arbitrarily fix $N$ pairs, $(x_1, t_1), \dots, (x_N, t_N)$, of an input and a threshold value.
  We upper bound the number of all possible tuples of $N$ Booleans (or outcomes) returned by the $N$ GJ algorithms, $\Gamma_{x_1, t_1},\dots,\Gamma_{x_N, t_N}$, whose input variable $\rho \in \R^\nparam $ is any $\nu$-sparse vector.
  By the definition of the pseudo-dimension (see \cref{def:dim}), we need at least $2^N$ outcomes to shatter $\set{x_1,\dots,x_N}$, and thus the largest such $N$ gives an upper bound on $\pdim(\Lcal)$.

  First, we fix a sparsity pattern $J \subseteq [\nparam ]$ with $|J| = \nu$ and let
  \[
    \Lcal_J = \Set*{L_\rho: \Xcal \to \R}{\rho \in \R^\nparam ,\ \supp(\rho) \subseteq J}.
  \]
  Note that we have $\Lcal = \bigcup_{J\subseteq [\nparam ]: |J| = \nu}\Lcal_J$.
  From the statement assumption, there is a $(\Delta, \pcomp)$-GJ algorithm $\Gamma_{x,t}$ that can check whether $L_\rho(x) > t$ or not.
  That is, for any $(x,t)$, whether $L_\rho(x) > t$ or not is determined by sign patterns of $\pcomp$ polynomials of degree at most $\Delta$ in $\rho \in \R^\nparam$.
  Moreover, since $\supp(\rho) \subseteq J$, $\Gamma_{x ,t}$ takes up to $\nu$ variables as input.
  Thus, once $J$ is fixed, outcomes of $\Gamma_{x_1, t_1},\dots,\Gamma_{x_N, t_N}$ are determined by sign patterns of $N\pcomp$ polynomials of degree at most $\Delta$ in $\nu$ variables.
  The number of such sign patterns is at most ${(8\e N\pcomp\Delta/\nu)}^\nu$ by Warren's theorem (\cref{thm:warren}).

  Next, we consider changing sparsity patterns.
  As discussed above, a fixed sparsity pattern $J$ yields up to ${(8\e N\pcomp\Delta/\nu)}^\nu$ outcomes of $\Gamma_{x_1, t_1},\dots,\Gamma_{x_N, t_N}$.
  If we feed $\rho \in \R^\nparam$ with a new sparsity pattern $J^\prime$ of size $\nu$ to $\Gamma_{x_1, t_1},\dots,\Gamma_{x_N, t_N}$, then $N\pcomp$ polynomials that appear in the GJ algorithms may exhibit up to ${(8\e N\pcomp\Delta/\nu)}^\nu$ new sign patterns, which lead to at most that many new outcomes.  Thus, when the sparsity pattern of $\rho$ can be any size-$\nu$ subset of $[\nparam]$, the number of all possible outcomes of $\Gamma_{x_1, t_1},\dots,\Gamma_{x_N, t_N}$ is at most
  \[
    \text{``the number of sparsity patterns''} \times {(8\e N\pcomp\Delta/\nu)}^\nu.
  \]
  Since there are up to $\binom{\nparam }{\nu} \le \nparam ^\nu$ sparsity patterns, the number of all possible outcomes of  $\Gamma_{x_1,t_1},\dots,\Gamma_{x_N,t_N}$ is at most $(8\e \nparam  N\pcomp\Delta/\nu)^\nu$.
  In order for $\Lcal$ to shatter $\set{x_1,\dots,x_N}$,
  \[
    2^N \le (8\e \nparam  N\pcomp\Delta/\nu)^\nu \Leftrightarrow N \le \nu \log_2(8\e \nparam  N\pcomp\Delta/\nu)
  \]
  must hold.
  Since $\log_2y \le \frac23y$ for $y > 0$, the right-hand side is bounded from above as
  \[
    \nu\log_2(8\e \nparam \pcomp\Delta) + \nu\log_2(N/\nu) \le \nu\log_2(8\e \nparam \pcomp\Delta) + \frac23N.
  \]
  Rearranging the terms, we obtain $N \le 3\nu\log_2(8\e \nparam \pcomp\Delta)$, hence $\pdim(\Lcal) = \Ord(\nu \log(\nparam \pcomp\Delta))$.
\end{proof}

\subsection{Result on Learning-Based LRA}\label{subsec:sparsity-lra}
We now return to the LRA setting and discuss the pseudo-dimension bound for the case of changeable sparsity patterns.
In this setting, we have $\nparam  = mn$ and $\nu = ns$ since every sketching matrix $S$ is of size $m\times n$ and has up to $ns$ non-zeros.
Furthermore, from the discussion in \cref{subsec:bartlett-overview,sec:improved-upper}, for any input $A \in \Acal$ and threshold value $t \in \R$, we can check whether the proxy loss value, $\Lhat_\varepsilon(S, A)$, exceeds $t$ or not by using a $(\Delta, \pcomp)$-GJ algorithm with
\begin{align}
  \Delta = \Ord(mk\varepsilon^{-1}\log(d/\varepsilon))
  \ \
  \text{and}
  \ \
  \pcomp = m\cdot 2^{\Ord(k)}\cdot (d/k)^{3k}.
\end{align}
Thus, from \cref{thm:sparse-main}, for the class $\hat\Lcal_\varepsilon = \set{\Lhat_\varepsilon(S, \cdot)}_{S \in \Scal}$ of proxy loss functions where $\Scal$ consists of sketching matrices with $ns$ non-zeros at any positions, it holds that
\[
  \pdim(\hat\Lcal_\varepsilon) = \Ord(ns \cdot (\log(mn) + k\log(d/k) + \log(1/\varepsilon))).
\]
The right-hand side is larger than the bound in \cref{prop:fat-dim-improved} only by $\Ord(ns\log n)$.
Note that narrowing the class $\Scal$ only decreases $\pdim(\hat\Lcal_\varepsilon)$; hence, the bound remains true when each $S \in \Scal$ is restricted to have $s$ non-zeros in each column.
Since we have $\eFSdim(\Lcal) \le \pdim(\hat\Lcal_\varepsilon)$ as discussed in \cref{subsec:bartlett-overview}, we obtain the following result.
\begin{proposition}\label{prop:sparsity-lra}
  Let $\Lcal = \set{L(S, \cdot)}_{S \in \Scal}$ be the class of loss functions defined by \eqref{eq:scw-loss} where $\Scal$ contains sketching matrices with any sparsity patterns of size $ns$.
  For sufficiently small $\varepsilon > 0$, the $\varepsilon$-fat shattering dimension of $\Lcal$ is bounded as
  \[
    \eFSdim(\Lcal) = \Ord(ns \cdot (\log(mn) + k\log(d/k) + \log(1/\varepsilon))).
\]
\end{proposition}


\section{EXPERIMENTS}\label{sec:experiment}
We confirm that learning sparsity patterns can improve the empirical accuracy of learning-based LRA methods.
Note that the uniform bound discussed in \cref{subsec:learning-theory} is agnostic to learning methods; therefore, we can use \cref{prop:sparsity-lra} to obtain generalization bounds for any methods to learn sparse sketching matrices.

\subsection{Background and Learning Methods}\label{subsec:experiment-background}
Let us first overview existing methods for learning sketching matrices.
\citet{Indyk2019-cn} initiated the study of learning-based LRA, as mentioned in \cref{subsec:lra}.
Assuming fixed sparsity patterns, they learned sketching matrices by applying SGD to the SCW-based loss \eqref{eq:scw-loss}, where gradients are computed via backpropagation through differentiable SVD.
\citet{Liu2020-hu} enhanced the previous method by first learning sparsity patterns with a greedy algorithm and then learning non-zeros via SGD.
A drawback of those two methods is that backpropagating through SVD is computationally expensive.
\citet{Indyk2021-yn} has overcome this issue by developing an efficient learning method based on a surrogate loss function.
While their method again assumes fixed sparsity patterns, we can naturally extend it to changeable sparsity patterns, as detailed later.
Another related work is \citep{Ailon2021-ia}, which proposed to represent linear layers of neural networks as products of sparse matrices, like the butterfly networks.
Although their idea is applicable to LRA, it requires sketching matrices with complicated structures; thus, we below do not consider it for simplicity.

Given the above background, an natural next direction is to extend the efficient method of \citep{Indyk2021-yn} to changeable sparsity patterns.
In \citep{Indyk2021-yn}, two kinds of methods are studied, one-shot and few-shot methods.
We focus on the latter and present how to modify it to learn both positions and values of non-zeros.
Their basic idea is to minimize the following surrogate loss instead of the SCW-based loss \eqref{eq:scw-loss}:
\begin{equation}\label{eq:surrogate-loss}
  \Ltl(S, A) = \norm{U_k^\top S^\top S U - I_0}_F^2,
\end{equation}
where $U \in \R^{n \times d}$ is the column orthogonal matrix computed by SVD of $A$ (assuming $\rank(A) = d$), $U_k \in \R^{n \times k}$ is the first $k$ columns of $U$ corresponding to the largest $k$ singular values, and $I_0 = [I_k, \zeros_{k, d-k}] \in \R^{k \times d}$ is a concatenation of the $k\times k$ identity matrix and $k \times d-k$ zeros.
Unlike the SCW-loss, differentiating the surrogate loss, $\Ltl(S, A)$, with respect to $S$ does not require backpropagation through SVD, hence more efficient.
Moreover, \citep[Theorem 2.2]{Indyk2021-yn} ensures the consistency of the surrogate loss, i.e., $\Ltl(S, A) \le \varepsilon$ implies $\norm{A - \scw_k(S, A)}_F^2 \le (1 + \Ord(\varepsilon)) \norm{A - \brc{A}_k}_F^2$.
By minimizing the empirical surrogate loss via SGD, they learned non-zeros of sketching matrices at fixed positions.

To learn both positions and values of non-zeros based on the above idea, we use the projected gradient descent method,
or sometimes called iterative hard thresholding (IHT) in non-convex sparse optimization \citep{Jain2017-wl}.
The method works iteratively as with SGD.
In each iteration, given an input matrix $A \in \Acal_\train$ in a training dataset, we update the sketching matrix as $S \gets \Pi_{s}(S - \eta \nabla \Ltl(S, A))$, where $\eta>0$ is a step size, $\nabla \Ltl(S, A)$ is the gradient with respect to $S$, and $\Pi_s$ is a projection operator that preserves the largest $s$ elements in absolute value for each column and set the others to zero.
In the following experiments, we refer to this method, which learns positions of non-zeros, as \learn and compare it with two baselines: \fix and \dense.
\fix is the method studied in \citep{Indyk2021-yn}, which learns non-zeros at fixed positions via SGD.
\dense learns values of all entries via SGD.
Note that although \dense naturally attains the best accuracy among them, it results in dense sketching matrices, which cannot benefit from the efficiency of sparse matrix multiplication and cause longer runtime of $\scw_k$ when deployed for future data.

\begin{figure*}[t]
	\centering
	\begin{minipage}[t]{.32\textwidth}
		\includegraphics[width=1.0\textwidth]{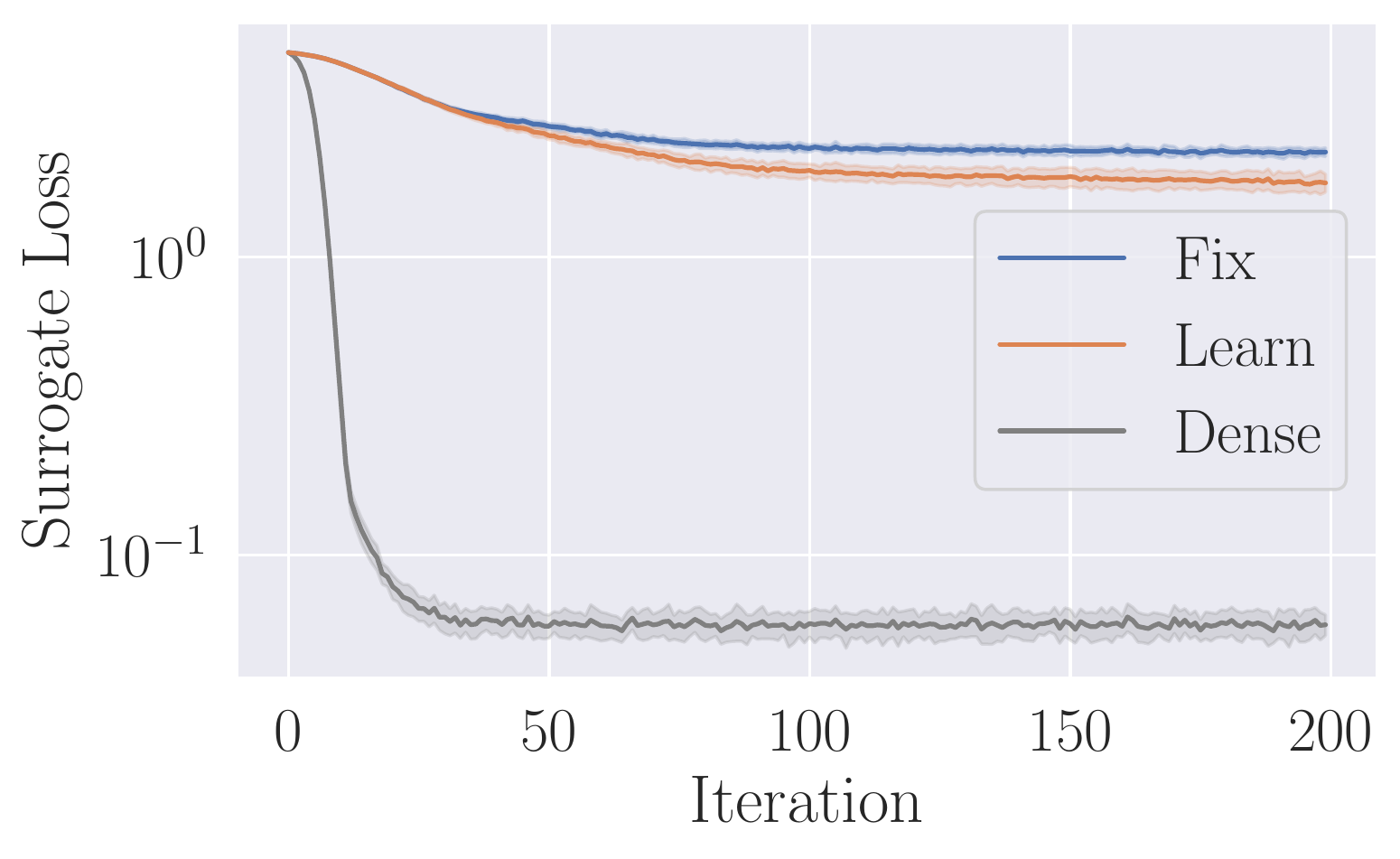}
		\subcaption*{Surrogate loss, $s=1$}
	\end{minipage}
	\centering
	\begin{minipage}[t]{.32\textwidth}
		\includegraphics[width=1.0\textwidth]{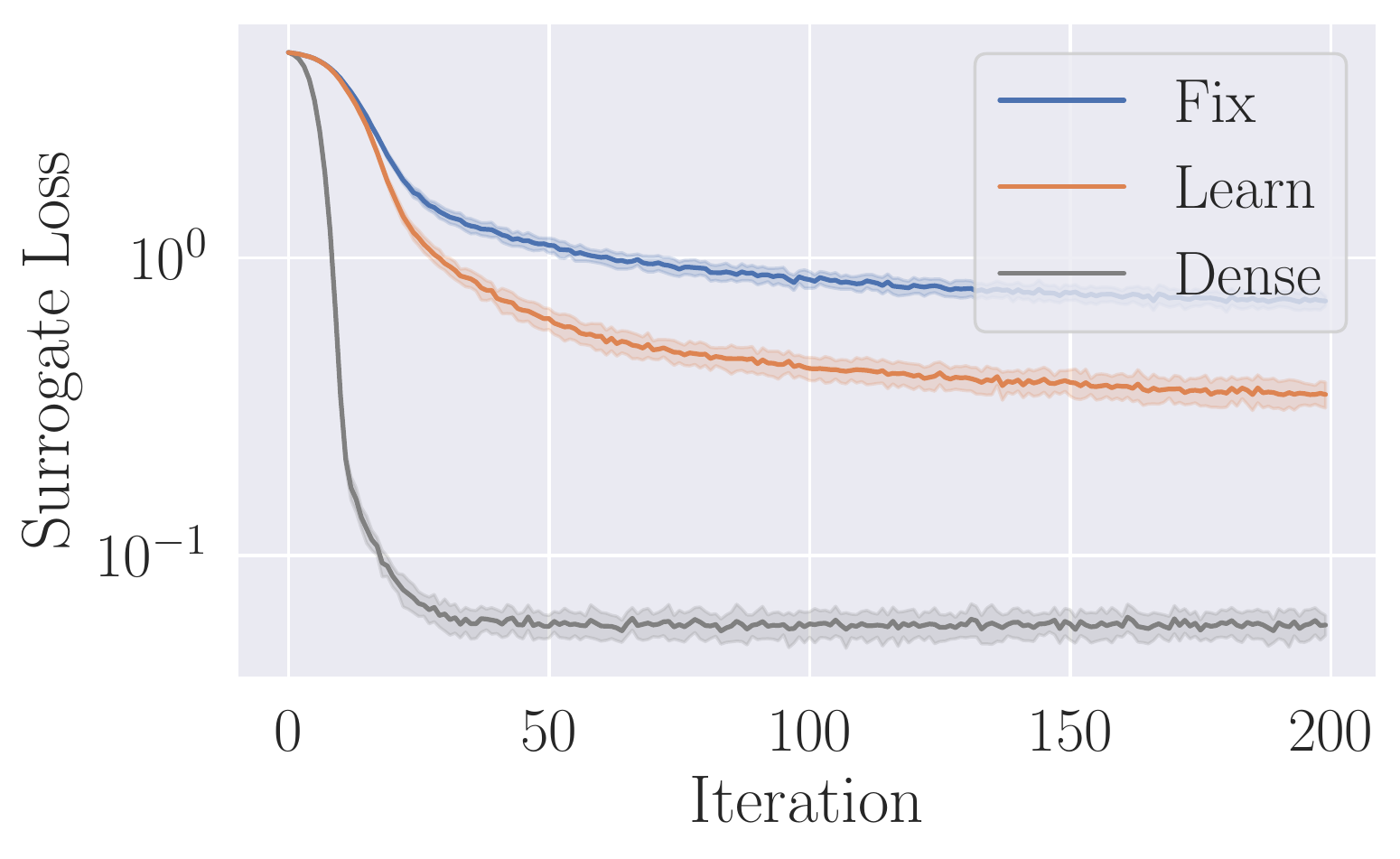}
		\subcaption*{Surrogate loss, $s=3$}
	\end{minipage}
	\centering
	\begin{minipage}[t]{.32\textwidth}
		\includegraphics[width=1.0\textwidth]{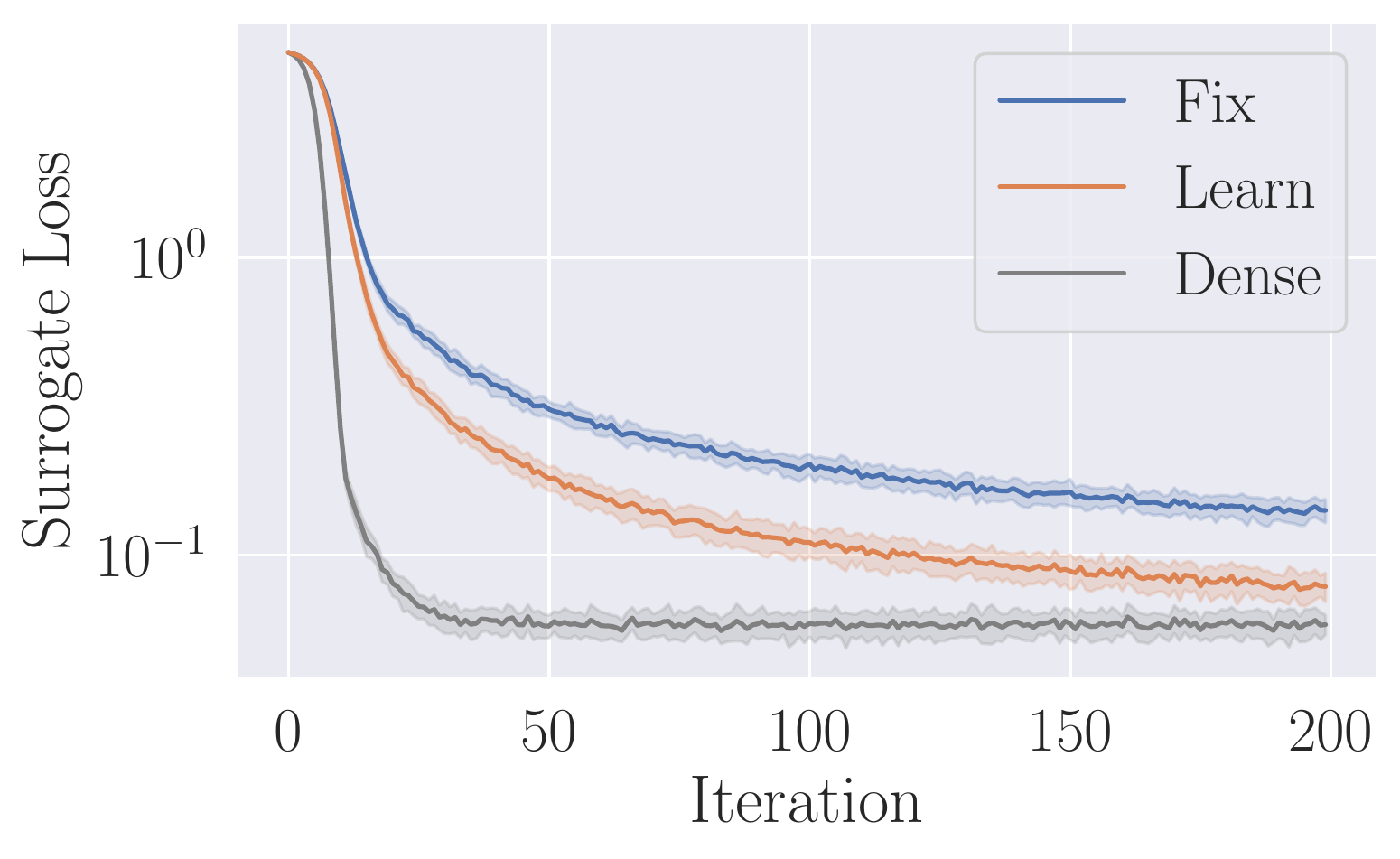}
		\subcaption*{Surrogate loss, $s=5$}
	\end{minipage}
	\centering
	\begin{minipage}[t]{.32\textwidth}
		\includegraphics[width=1.0\textwidth]{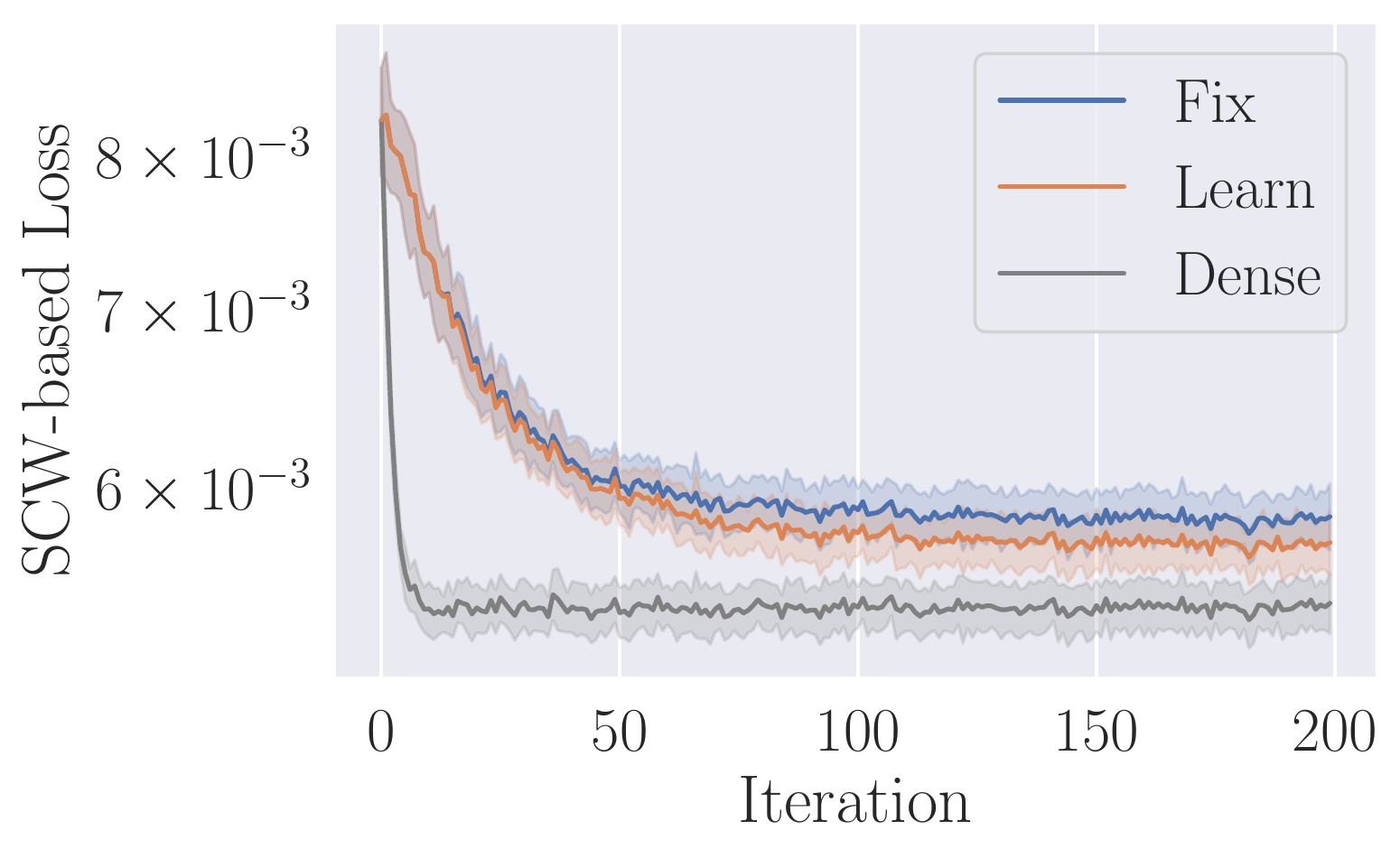}
		\subcaption*{SCW loss, $s=1$}
	\end{minipage}
	\centering
	\begin{minipage}[t]{.32\textwidth}
		\includegraphics[width=1.0\textwidth]{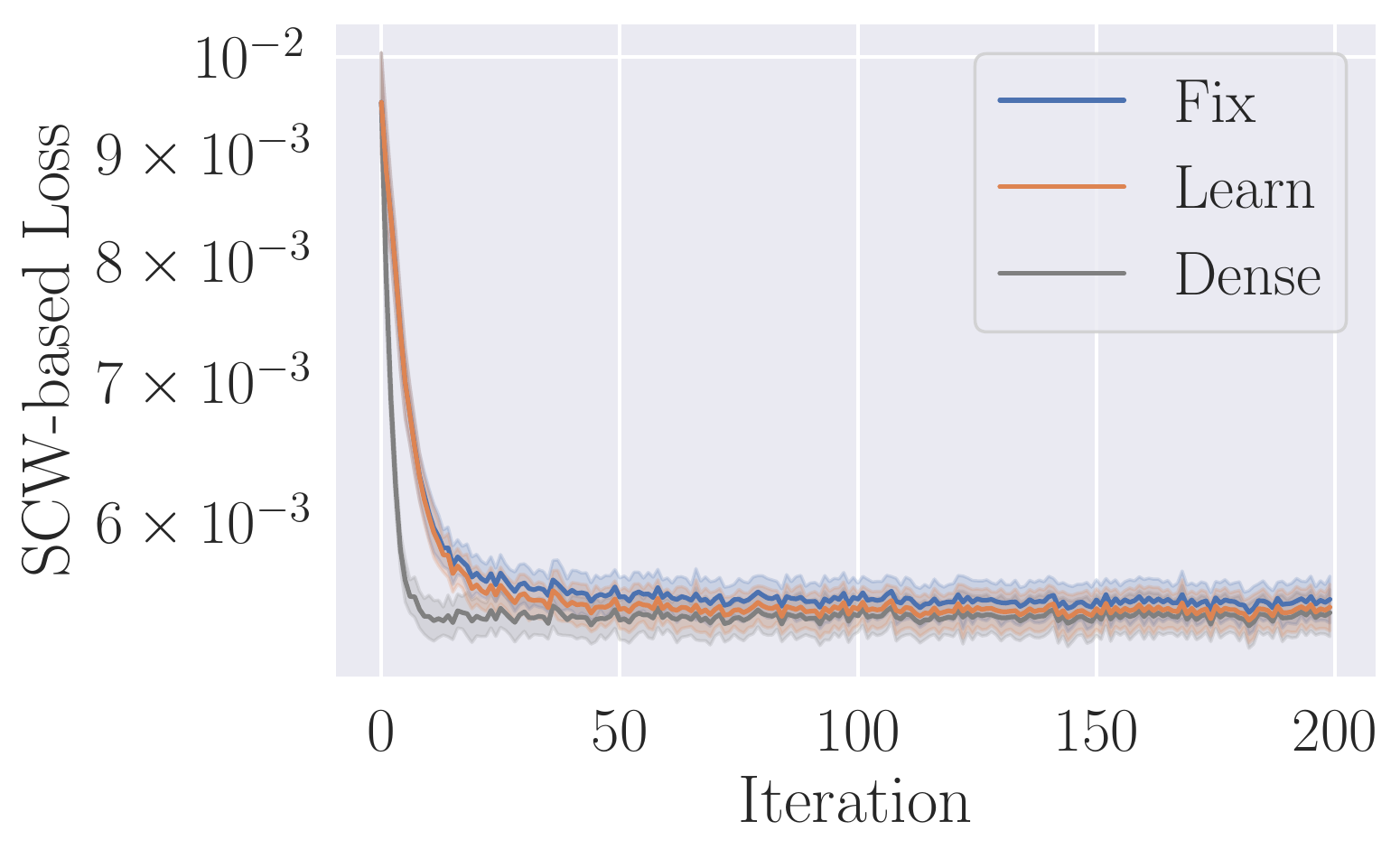}
		\subcaption*{SCW loss, $s=3$}
	\end{minipage}
  \centering
	\begin{minipage}[t]{.32\textwidth}
		\includegraphics[width=1.0\textwidth]{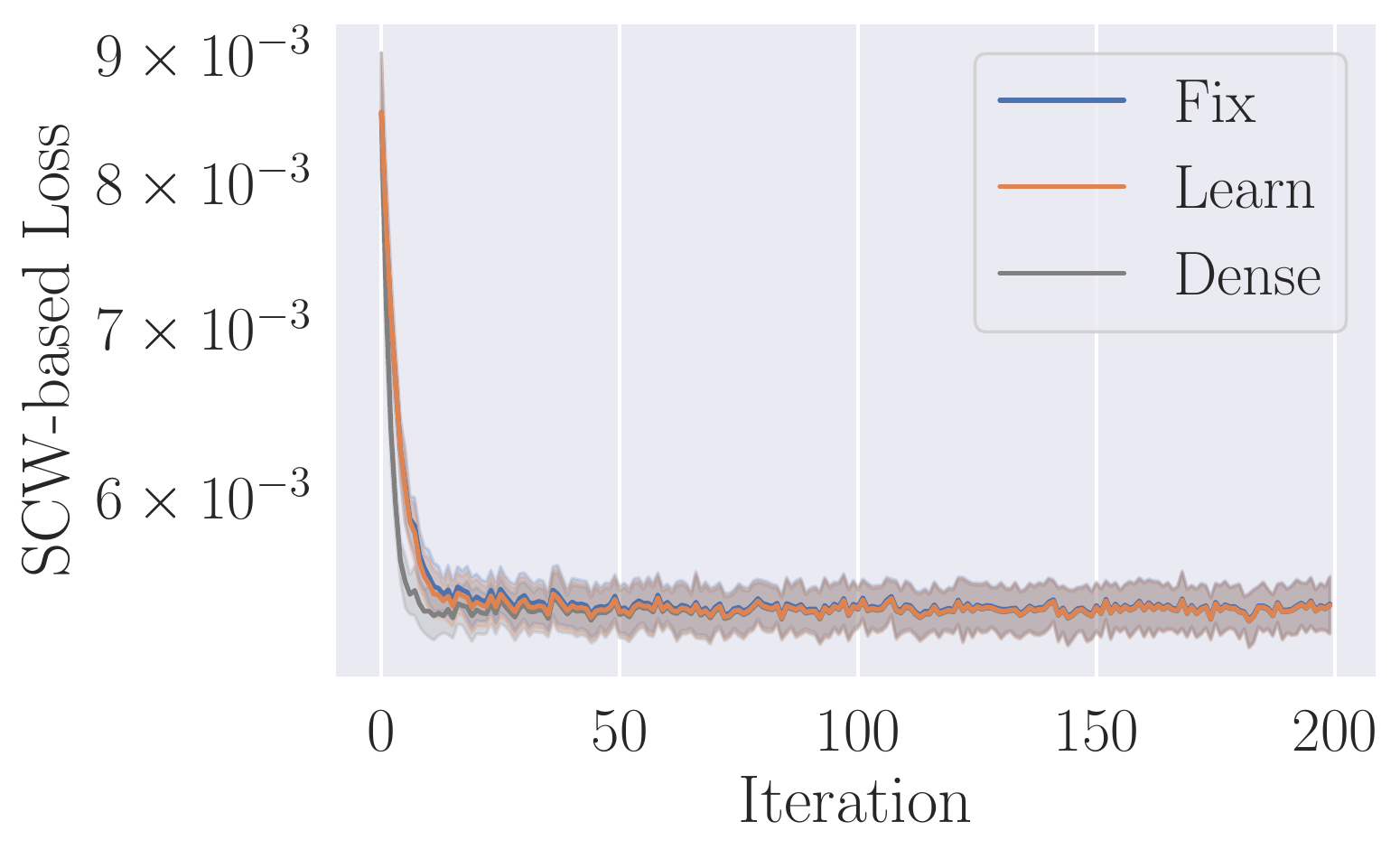}
		\subcaption*{SCW loss, $s=5$}
	\end{minipage}
  \caption{
    Surrogate and SCW-based loss values on training datasets.
    The x-axis indicates the number of iterations of SGD (for \fix and \dense) or stochastic IHT (for \learn).
    The error band indicates the standard deviation over the $30$ random trials.
  }
  \label{fig:train}
\end{figure*}

\begin{figure}[t]
	\centering
  \includegraphics[width=.45\textwidth]{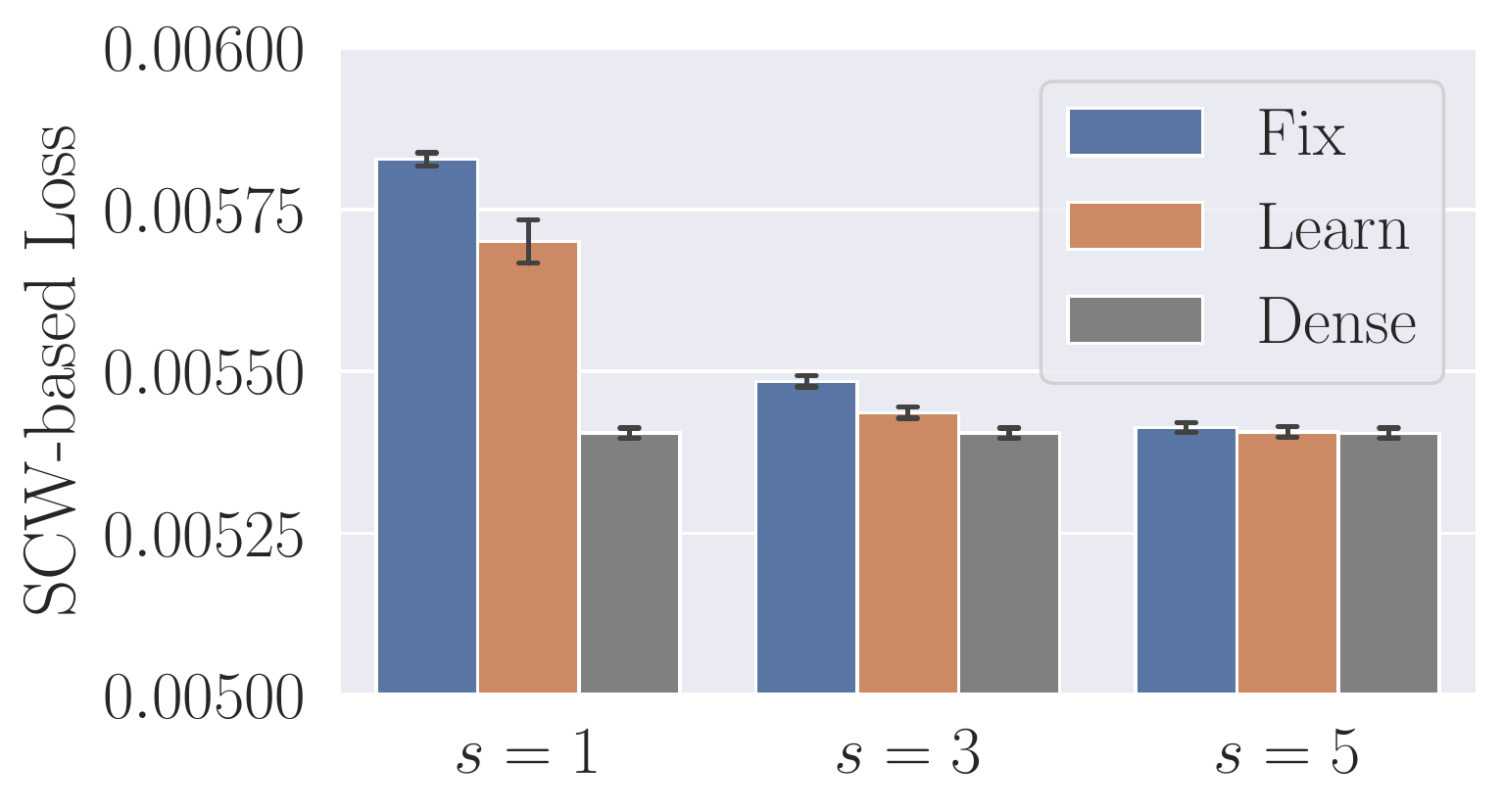}
  \caption{
    SCW-based loss values on test datasets. The error bar shows the standard deviation over the $30$ random trials.
  }
  \label{fig:test}
\end{figure}

\subsection{Settings and Results}
Experiments were conducted on a macOS machine with Apple M2 CPU and 24 GB RAM.
We implemented the methods in Python 3.9.12 and used JAX 0.3.15 \citep{Bradbury2018-xq} to compute gradients.
When performing SVD, we regarded singular values smaller than $10^{-8}$ as zero.

Let $n = 100$, $d = 50$, $m = 10$, and $k = 5$.
We made a rank-$k$ matrix $A_\text{true} \in \R^{n \times d}$ by multiplying $n\times k$ and $k\times d$ matrices whose entries were drawn from the uniform distributions over $[0, 1]$.
We then let $A = A_\text{true} + 0.1\times A_\text{noise}$, where entries of $A_\text{noise}$ were drawn from the standard normal distributions, and normalized $A$ so that $\norm{A}_F = 1$ holds.
By drawing $300$ noise terms independently, we created a dataset of $300$ input matrices $A$.
We split them into training and test datasets of sizes $200$ and $100$, respectively.
We made $30$ random training/test splits to calculate the average and standard deviation over the $30$ random trials.

We learn sketching matrices $S \in \R^{m \times n}$ by minimizing the empirical surrogate loss \eqref{eq:surrogate-loss} on a training dataset.
\fix and \learn learn $S$ with $s=1$, $3$, or $5$ non-zeros in each column; since $m=10$, the $s$ values mean that $10\%$, $30\%$, or $50\%$ of entries can be non-zero, respectively.
Initial sketching matrices were obtained by setting random $s$ entries in each column to $-1$ or $+1$ with probability $0.5$, respectively, and the others to zero; we then normalized it to satisfy $\norm{S}_F = 1$ for numerical stability.
We set the step size, $\eta$, to $0.1$.

\cref{fig:train} shows curves of surrogate \eqref{eq:surrogate-loss} and SCW-based \eqref{eq:scw-loss} loss values in the training phase.
As $s$ increased, the performances of \fix and \learn became closer to that of \dense.
Regarding the surrogate loss, \learn achieved smaller values than \fix, implying that \learn could go beyond local optima into which \fix fell.
As for the SCW-based loss, \learn slightly outperformed \fix for $s=1$ and $3$, and both achieved almost as small values as \dense when $s=5$.

\cref{fig:test} shows the SCW-based loss values on test datasets.
As with the training SCW-based loss values (\cref{fig:train}), the gap between \fix and \learn was evident with $s=1$ and $3$, while both achieved as small losses as \dense with $s=5$.

To conclude, \learn achieved smaller SCW-based loss values than \fix particularly when $s$ was small, suggesting that learning sparsity patterns enables more accurate learning-based LRA when we need to learn highly sparse sketching matrices for the sake of the efficiency of $\scw_k$.

As for training times, \learn took about $8\%$ longer than \fix, although our main focus is accuracy and the implementations are not intended to be fast.

\section{CONCLUSION AND DISCUSSION}
Building on \citep{Bartlett2022-mu}, we have studied generalization bounds for learning-based LRA.
We have improved their $\tilde\Ord(nsm)$ bound on the fat shattering dimension to $\tilde\Ord(nsk)$ by developing an $(\Ord(m), m)$-GJ algorithm that computes a pseudo-inverse of a matrix with $m$ rows.
We have also demonstrated its usefulness by applying it to the learning-based Nystr\"om method.
Then, we have shown that learning both positions and values of non-zeros of sketching matrices increases the fat-shattering-dimension bound only by $\Ord(ns\log n)$.
Experiments have confirmed that the efficient learning method of \citep{Indyk2021-yn} can achieve higher empirical accuracy with changeable sparsity patterns.

A notable open problem is to close the $\tilde\Ord(k)$ gap between the $\tilde\Ord(nsk)$ upper and $\Omega(ns)$ lower bounds.
Note that only applying our GJ algorithm to \cref{item:choose-Z} in \cref{def:proxy} does not leave out the $\tilde\Ord(k)$ factor;
a more essential problem lies in \cref{item:for-all-Pi}, where we must avoid using exponentially many $P_i$ in $k$ to remove the $\tilde\Ord(k)$ factor.
When it comes to improving the $\Omega(ns)$ lower bound, we need to shatter more instances than $\Omega(ns)$, where $ns$ is the number of tunable parameters.
Although obtaining a greater lower bound than the number of tunable parameters is typically challenging, such lower bounds have been obtained for neural networks using the \emph{bit extraction} technique \citep{Bartlett1998-qs}.
We expect that a similar idea would help obtain a tighter lower bound.

\section*{Acknowledgements}
This work was supported by JST ERATO Grant Number JPMJER1903 and JSPS KAKENHI Grant Number JP22K17853.

\bibliographystyle{abbrvnat}
\bibliography{DataDrivenLinearAlgebra}


\clearpage
\appendix

\thispagestyle{empty}

\end{document}